\documentclass[conference]{IEEEtran}
\IEEEoverridecommandlockouts
\usepackage{cite}
\usepackage{amsmath,amssymb,amsfonts}
\usepackage{graphicx}
\usepackage{textcomp}
\usepackage{xcolor}
\def\BibTeX{{\rm B\kern-.05em{\sc i\kern-.025em b}\kern-.08em
    T\kern-.1667em\lower.7ex\hbox{E}\kern-.125emX}}

\usepackage{amsmath}
\usepackage{mathtools}
\usepackage{amssymb}
\usepackage[acronym,toc]{glossaries}
\usepackage[noend]{algpseudocode}
\usepackage{algorithm}
\newcommand{\bigO}{\mathcal{O}}

\newacronym{ml}{ML}{machine learning}
\newacronym{dl}{DL}{Deep learning}
\newacronym{ai}{AI}{artificial intelligence}
\newacronym{cnn}{CNN}{convolutional neural network}
\newacronym{nn}{NN}{neural network}
\newacronym{relu}{ReLU}{rectified linear unit}
\newacronym{fc}{FC}{fully-connected}
\newacronym{knn}{k-NN}{k-nearest neighbours}
\newacronym{svm}{SVM}{support vector machine}
\newacronym{hog}{HoG}{histograms of oriented gradients}
\newacronym{sift}{SIFT}{scale invariant feature transform}
\newacronym{csir}{CSIR}{Council for Scientific and Industrial Research}
\newacronym{mds}{MDS}{Modelling and Digital Science}
\newacronym{mias}{MIAS}{Mobile Intelligent Autonomous Systems}
\newacronym{sgd}{SGD}{stochastic gradient descent}
\newacronym{vgg}{VGG}{Visual Geometry Group}
\newacronym{eda}{EDA}{exploratory data analysis}
\newacronym{tsne}{t-SNE}{T-distributed stochastic neighbour embedding}
\newacronym{cam}{CAM}{class activation maps}
\newacronym{gp}{GP}{Gaussian process}
\newacronym{dli}{DLI}{Deep Learning Indaba}
\newacronym{wiml}{WiML}{Women in Machine Learning}
\newacronym{bai}{BAI}{Black in AI}
\newacronym{rl}{R0L}{reinforcement learning}
\newacronym{nlp}{NLP}{natural language processing}
\newacronym{pdf}{PDF}{probability density function}
\newacronym{ei}{EI}{expected improvement}
\newacronym{lcb}{LCB}{Lower Confidence Bound}
\newacronym{mse}{MSE}{mean squared error}
\newacronym{se}{SE}{squared exponential}
\newacronym{pi}{PI}{probability of improvement}
\newacronym{cpf}{CPF}{cumulative probability function}
\begin{document}

\title{Black-Box Saliency Map Generation Using Bayesian Optimisation}

\author{\IEEEauthorblockN{Mamuku Mokuwe}
\IEEEauthorblockA{\textit{Centre for Robotics and Future Production} \\
\textit{Council for Scientific and Industrial Research}\\
Pretoria, South Africa \\
mmokuwe@csir.co.za}
\and
\IEEEauthorblockN{Michael Burke}
\IEEEauthorblockA{\textit{School of Informatics} \\
\textit{University of Edinburgh}\\
Edinburgh, Scotland \\
Michael.Burke@ed.ac.uk}
\and
\IEEEauthorblockN{Anna Sergeevna Bosman}
\IEEEauthorblockA{\textit{Department of Computer Science} \\
\textit{University of Pretoria}\\
Pretoria, South Africa \\
anna.bosman@up.ac.za}
}
\maketitle


\begin{abstract}
Saliency maps are often used in computer vision to provide intuitive interpretations of what input regions a model has used to produce a specific prediction. A number of approaches to saliency map generation are available, but most require access to model parameters. This work proposes an approach for saliency map generation for black-box models, where no access to model parameters is available, using a Bayesian optimisation sampling method. The approach aims to find the global salient image region responsible for a particular (black-box) model's prediction. This is achieved by a sampling-based approach to model perturbations that seeks to localise salient regions of an image to the black-box model. Results show that the proposed approach to saliency map generation outperforms grid-based perturbation approaches, and performs similarly to gradient-based approaches which require access to model parameters.
\end{abstract}

\section{Introduction}

\gls{dl} techniques have become a standard approach in computer vision. Specifically, the \gls{cnn} architecture has shown exceptional performance, achieving results comparable to human performance on image recognition tasks \cite{ref:Krizhevsky:2012:ICD, ref:Simonyan:2014, ref:Szegedy:2015}. As a result, the \gls{cnn} models are often deployed in real life as efficient black-box tools. However, there remains a gap in our ability to explain and/or interrogate a model's decisions. Saliency visualisation in \gls{ml} is a type of visualisation that provides an intuitive explanation of the model's output by highlighting the input regions which contributed the most to the final output. The key motivation behind saliency visualisation in computer vision is to be able to distinguish the image region that contains the information responsible for the model's prediction \cite{ref:ittit:2000, ref:Simonyan:2013, ref:Zeiler:2013, ref:Baehrens:2009c}. By providing interpretable details about the model's decision, end-user trust 
can be gained, and introspection can be performed when predictions are incorrect.

Most existing approaches for saliency map generation assume that the model of choice is a white-box, i.e., that certain characteristics of the model are known, and that access to the model's parameters is available \cite{ref:Smilkov:2017, ref:Springenberg:2014}. In this work, we introduce a saliency map generation approach that can be used for any black-box model, only requiring access to input and output data, and the ability to query the model. This is particularly important for introspection into third-party tools, or the investigation of models where access to the underlying parameters is not available.

\begin{figure}[!bt]
    \centering
    \includegraphics[width=0.47\textwidth]{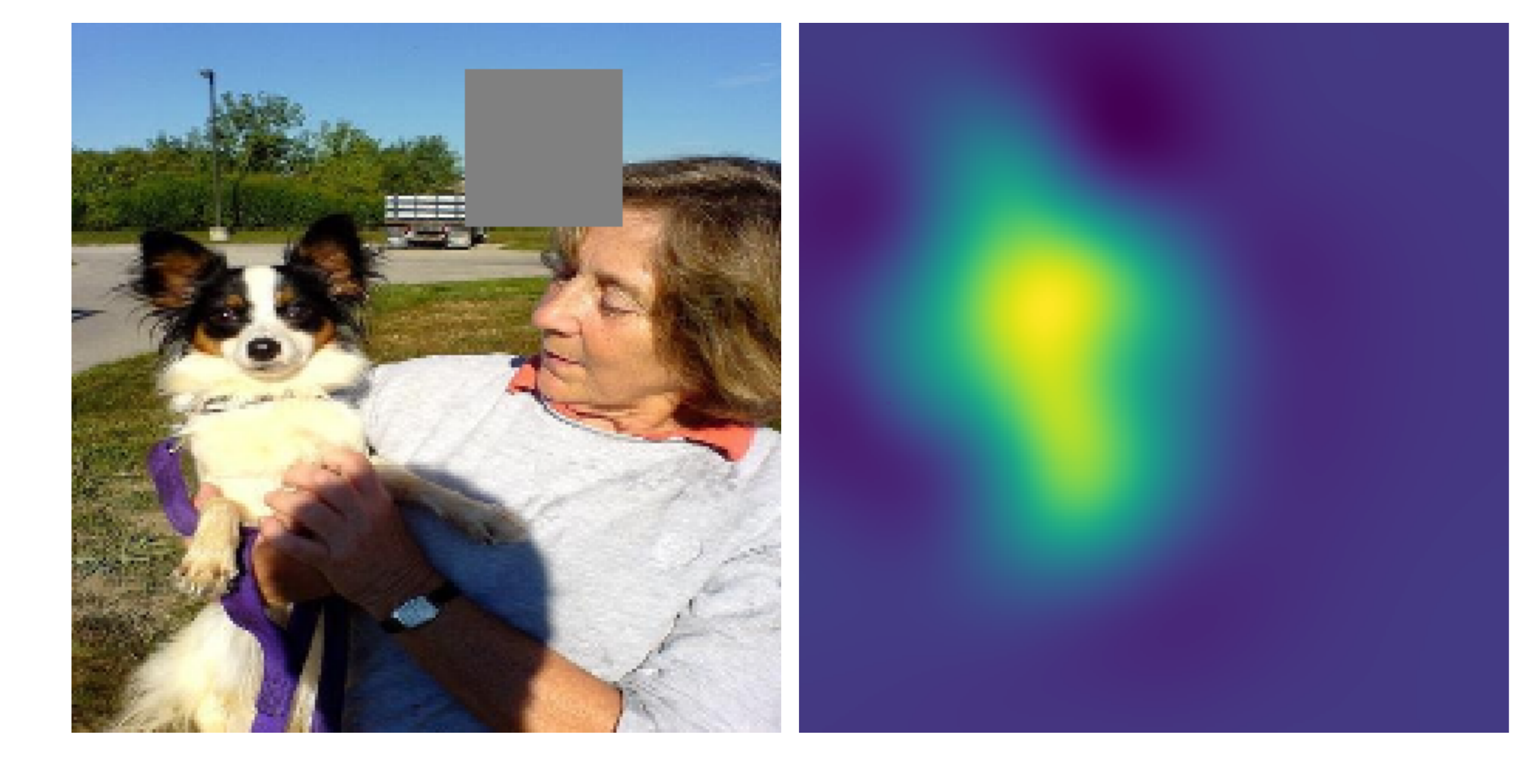}
    \caption{Example of the saliency map (right) and the corresponding input image (left) for class-prediction ``papillon''. The input image has an overlayed window box, representing an area used to measure sensitivity of the image location to the model's output.}
    \label{fig:fig6}
\end{figure}

The proposed approach generalises a standard occlusion-based sliding window technique for saliency map generation~\cite{ref:Zeiler:2013}. The standard occlusion-based approach (referred to as exhaustive search further in this study) involves sequentially blanking regions of the input, and measuring the resultant change in model output. The intuition behind the blanking operation is that for a given model $f$, an image $X$, and a a partially blanked image $\hat{X}$, the model outputs $f(X)$ and $f(\hat{X})$ should vary significantly if an important feature of $X$ was blanked in $\hat{X}$. An example of a partially blanked image and the corresponding saliency map is shown in Figure~\ref{fig:fig6}. Using the exhaustive search approach requires many blanking operations, making it computationally expensive, so typically only very coarse saliency maps can be generated. An important parameter for the exhaustive search is the size of the blanking window. Multiple window sizes can be employed to improve the quality of the saliency map, however, every additional window size further reduces the computational efficiency of the approach.

A method proposed by Burke~\cite{ref:Burke:2017} assumes that the \gls{ml} model is a black box, and uses probabilistic inference to mimic the behaviour of the model in response to small changes made to the input image. The sampling is performed in areas of interest to produce the saliency maps. This approach reduces the need to sample at every pixel location, however, there is still a need to manually specify the size of the blanking window used to occlude image regions.

This study builds on the work of Burke~\cite{ref:Burke:2017} to find a general occlusion-based saliency method for black-box models. A Bayesian optimisation sampling approach is used to look for potential regions to blank in an image. Additionally, Bayesian optimisation is used to identify the best blanking window size to use per sample. As a result, the salient parts of the image are sampled more densely, thus reducing the computational cost of producing a saliency map. The primary contributions of this work are as follows:
\begin{itemize}
\item We introduce a Bayesian optimisation approach to occlusion-based saliency map generation for black-box models.
\item We propose an intersection over union saliency metric to quantitatively measure the accuracy of a saliency map.
\end{itemize}

The rest of the paper is structured as follows: Section~\ref{sec:bg} provides the background on saliency map generation. Section~\ref{sec:occlusion} gives a formal definition of the occlusion-based saliency map generation method. Sections~\ref{sec:bayes} discusses the proposed Bayesian optimisation approach to saliency map generation. Section~\ref{sec:experiments} describes the experimental procedure, and presents the empirical results. Section~\ref{sec:conclusion} concludes the paper.

\section{Background}\label{sec:bg}

Early work on saliency map generation algorithms~\cite{ref:Baehrens:2009c, ref:Simonyan:2013, ref:Zeiler:2013} aimed to calculate the gradients from the model to identify pixels that contributed more to the model's output. The derived gradient was accessed using a single pass through the model, and did not take the model architecture, eg. individual network layers, into account. Fong and Vedaldi~\cite{ref:Fong:2018} proposed a method that builds on \cite{ref:Simonyan:2013}. Rather than using one pass through the model to obtain the gradient used for the saliency map formation, Fong and Vedaldi's method derives the gradient multiple times using the gradient descent optimisation technique. At each iteration, more information is learned that helps find the size of the perturbation mask which minimises the image classification error. In the reported results, $300$ iterations were used to produce the saliency map~\cite{ref:Fong:2018}. This approach is not applicable to true, non-differentiable black-box models, and finite difference gradient approximations may be expensive.

More recent approaches \cite{ref:Zhou:2015, ref:Selvaraju:2016, ref:Springenberg:2014} have used gradient-based techniques that explicitly manipulate gradients at different network layers. Zhou et al.~\cite{ref:Zhou:2015} proposed a saliency mapping approach that requires a direct extraction of the learned weights from the last convolutional layer of a \gls{cnn}, and applies global average pooling~\cite{ref:Lin:2013} to generate \gls{cam}. However, this method can only be used for image classification tasks; thus, a generalisation of \gls{cam} was proposed by Selvaraju et al.~\cite{ref:Selvaraju:2016} to extend the approach to accommodate any \gls{cnn}. Even though these techniques generate saliency maps that are interpretable, they require direct access to the learned features of a \gls{cnn}, and are designed for the specific \gls{cnn}-type architecture. 

Dabkowski and Gal~\cite{ref:Dabkowski:2017} propose a method that extracts feature maps at multiple layers and optimises the regions accountable for the output such that (1) the salient region alone allows a sufficient confidence classification, and (2) when the saliency region is removed, the model is not able to predict the correct class of the image. These objectives are further used to define a saliency metric that seeks to identify if the saliency map has indeed discovered the most discriminant region that contributed to a model's prediction. 

All of the approaches discussed above require access to model parameters, which makes them unsuitable for black-box model introspection.

A general approach for saliency map generation for black-box models requires testing the model's sensitivity through small alterations to the input, as demonstrated by Zeiler and Fergus~\cite{ref:Zeiler:2013}. This process sequentially blanks parts of the image region to find input regions that maximally contribute to the final prediction. A variation of this approach was proposed by Burke~\cite{ref:Burke:2017}. The saliency overlay for black-box models proposed in~\cite{ref:Burke:2017} uses a \gls{gp} for saliency map approximation, with a random sampling strategy to search for potential areas to mask in the input image. A limitation of this method is the use of heuristics to select the blanking window size to be used for probing the model. This study proposes a modification of~\cite{ref:Burke:2017} that automatically chooses the appropriate blanking window size using Bayesian optimisation.

The next section provides a formal definition of the exhaustive occlusion-based method for saliency map generation. Subsequently, the proposed Bayesian optimisation approach is discussed in detail.
\section{Occlusion-based saliency}
\label{sec:occlusion}
Occlusion-based saliency map generation sequentially blanks patches of input pixels, and evaluates the model's output with respect to the blanked image location. This can be considered a crude form of sensitivity analysis that identifies salient regions in the image by measuring the change in model output, without tampering with the model.

Without loss of generality, the following definition assumes the model of interest to be a classification model. Let $\displaystyle{f}$ be the image classification model and $\displaystyle{X}$ be the images. To evaluate how well $\displaystyle{f}$ performs given $\displaystyle{X}$, we observe the relationship between the blanked image $\displaystyle{\hat{X}}$ and the change in the model's output, i.e. class probability, $\displaystyle{y = f(X) - f(\hat{X})}$. The saliency map is formed by the $\displaystyle{y}$ values, which are obtained by sequentially blanking the image per pixel location, and passing the blanked image $\displaystyle{\hat{X}}$ to the model.

The exhaustive per-pixel approach requires many model evaluations, where uninteresting regions are also blanked multiple times. This approach is typically infeasible for larger images, and a grid-based approximation is generally proposed for computational feasibility.

The next section presents the proposed Bayesian optimisation approach for saliency map generation.
\section{Bayesian optimisation for saliency generation}
\label{sec:bayes}

Bayesian optimisation is a search method that locates the global optima of a black-box function through sequential sampling. The sequential process alternates between two steps: (1) fitting a probabilistic model that approximates the global point (or region) of $f$, and (2) using a manipulation function to determine the next best point to evaluate based on the previous mean and variance predictions~\cite{ref:Brochu:2010a}. This work aims to exploit this method for the purpose of saliency map generation. The primary goal of this work is to generate the saliency map using the search method that chooses globally optimal sampling parameter values for the blanking process, such that the greatest reward is yielded for faster saliency map convergence.

\begin{figure}[!b]
    \centering
    \includegraphics[width=0.5\textwidth]{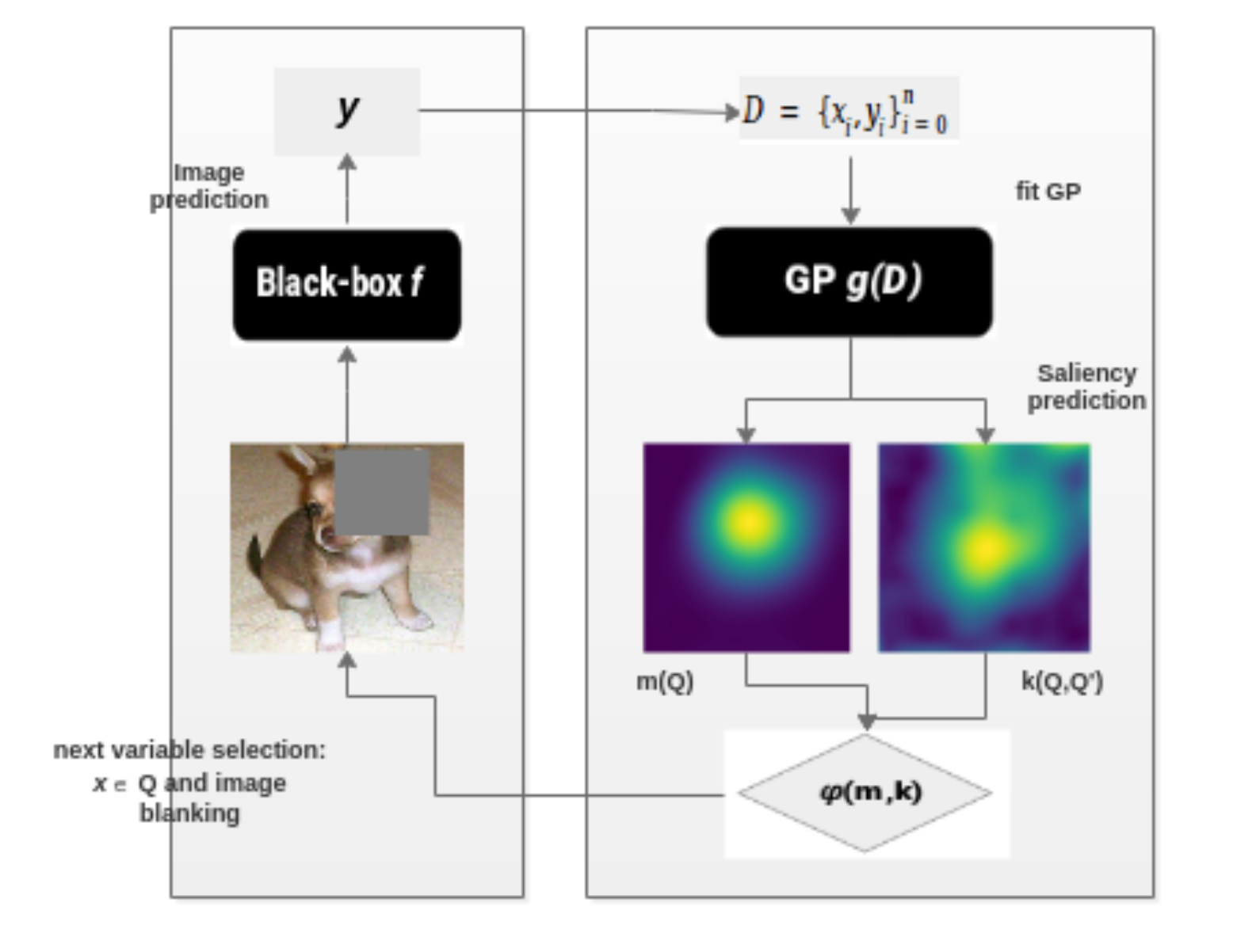}
	\caption{Bayesian optimisation process for the generation of a saliency map.}
	\label{fig:fig1}
\end{figure}

Figure \ref{fig:fig1} provides a diagrammatic overview of the proposed saliency map generation algorithm using Bayesian optimisation. The black-box model is accessed to extract the information used to generate the saliency map through Bayesian optimisation, which involves fitting a \gls{gp} model to the observations, and using an acquisition function to select the next region to sample, together with the window size. Section~\ref{sub:gauss} discusses how the \gls{gp} model is fit. Section~\ref{sub:aquisition} discusses the acquisition function employed in the study. Section~\ref{sub:eval} proposes a region-based ratio metric to be used for the performance evaluation of the saliency maps.

\subsection{Gaussian process}\label{sub:gauss}
Generating the saliency map for a given image requires iteratively fitting a \gls{gp} model to determine the posterior probability over functions. A \gls{gp} by definition is a collection of normally distributed random variables where any finite number of the variables have a jointly Gaussian distribution \cite{ref:Osborne:2009}. Intuitively, the \gls{gp} is a generalisation of the multivariate Gaussian distribution that is completely specified by the mean function $\displaystyle{m(Q)}$ and the covariance function $\displaystyle{k(Q,Q^{'})}$, given a set of observations $Q$. The \gls{gp} is used to construct a probabilistic model $g$ using any finite random collection of data, $\mathcal{D} = \{(\mathbf{x}_{1:i}, y_{1:i})\}; \hspace{2mm} \mathbf{x}_{1:i} = \{(u_{1:i},v_{1:i},s_{1:i})\}$ and $y_{1:i} = \{f(X)_{1:i} - f(\hat{X})_{1:i}\}$, which is expected to approximate the black-box function $f$ after sufficient observations have been made. Thus, fitting a \gls{gp} model $g$ requires a finite set of input and output observations from the black-box model $f$. Let $X$ be an $m \times n$ input image given by a set $\{(u_1,v_1), ..., (u_m,v_n)\}$  of all possible pixel coordinate combinations, and let $S = \{s_1, ..,s_l\}$ be the set of variable window sizes of length $l$. We can fit $g$ using a set of samples $D_{1:i} = \{\mathbf{x}_{1:i}, y_{1:i}\}$. The model $g$ can later be used to predict the approximated saliency map for other unseen observations $Q = \{\mathbf{x}_1, ..., \mathbf{x}_p\}$, $p = m\times n\times l$. 
 

To begin the algorithm, a random point(s) $\mathbf{x}_i \in Q$ is selected, the image $X$ is blanked at coordinates $(u_i,v_i)$ with window variable size $s_i$, and fed to the model to obtain the output $\displaystyle{f(\hat{X}_i)}$ before \gls{gp} evaluation. The finite set $\displaystyle{D_{1 : i} = \{\textbf{x}_{1 : i},y_{1 : i}\}}$, with variables added continuously during the optimisation process, is then used to build the prior \gls{gp} model $g$. For \gls{gp} model parameter choice, the Mat\'ern kernel function is selected. 
The kernel function is used to estimate the covariance $\displaystyle{k(Q,Q^{'})}$ of the Gaussian function. Genton~\cite{ref:Genton:2002} discusses different types of kernel functions for \gls{ml} and their impact. The Mat\'ern covariance function is a kernel function that assumes an uneven feature space. This property introduces flexibility that makes the Mat\'ern function applicable to practical problems \cite{ref:Stein:1999}. The following equation defines the Mat\'ern function:
\begin{equation}
    k_{mat}(\mathbf{x}_i, \mathbf{x}_j) = \frac{\sigma^22^{\nu -1}}{\Gamma (\nu)}\Big(\frac{\sqrt{2\nu}}{l}r\Big)^{\nu} k_{v}\Big(\frac{\sqrt{2\nu}}{l}r\Big),
    \label{eq:eq3}
\end{equation}
where $\sigma$ is the kernel variance, $r = |\mathbf{x}_i - \mathbf{x}_j|$, $\Gamma(*)$ is the gamma function, $k_{v}(*)$ is the Bessel function, and $\nu = 2.5$ and $l = 12$ are the hyperparameters of the Mat\'ern function, found using maximum likelihood estimation.


To fit the \gls{gp} iteratively, Bayes' theorem is used, which provides the updated expected solution based on the previously observed variables as follows:
\begin{eqnarray}
p(y | \mathbf{x}) = \frac{p(\mathbf{x} | y)p(y)}{p(\mathbf{x})}, \label{eq:eq6}
\end{eqnarray}
where $p(\mathbf{x} | y)$ is the likelihood, $p(\mathbf{x})$ is the evidence, and $p(y)$ is the prior. To improve the saliency map, we iteratively update the prior belief of $g$ by adding new observations to ${D}$. The new observations are selected such that they improve the model $g$'s knowledge about the true function $f$ or reduce the number of sample variables needed to generate the saliency overlay. To find these observations, we use an acquisition function that uses the posterior mean and covariance values to specify the next pixel coordinates of the image to sample at, together with the window size to blank. Aquisition functions are discussed in the next section.
\subsection{Acquisition functions}\label{sub:aquisition}
In order for the Bayesian optimisation saliency map approach to converge in minimal time, a clever way of selecting regions to blank in the input image that yields the highest reward must be followed. In the Bayesian optimisation approach, an acquisition function $\varphi$ is used to find the next optimal point $\mathbf{x}_i^{*} \in Q$ observations, using the predicted posterior mean and covariance \cite{ref:Osborne:2009}. 

The commonly studied acquisition functions for finding the global optimum of the black-box function can be found in \cite{ref:Brochu:2010a}. For this work, we used the \gls{ei} acquisition function. The \gls{ei} acquisition function implicitly balances the exploration-exploitation trade-off, thereby exploring places with high variance, while exploiting at places with low mean~\cite{ref:Garnett:2015}. The balance between the exploration and exploitation helps the \gls{gp} model gain more certainty of the black-box function as a whole.

As shown in Figure~\ref{fig:fig1}, the acquisition function is used to compute the next set of elements for the blanking process. The \gls{ei} acquisition function is given by:
    \begin{equation}\label{eq:eq10}
    \begin{split}
     \varphi_{EI} = (m(Q)_i - y_i^{*})\Phi(Z) - k(Q,Q^{'})_{i} \phi(Z),\\
     Z = \frac{m(Q)_i - y_i}{k(Q,Q^{'})_{i}},                     
     \end{split}
    \end{equation}
    where $\Phi(Z)$ and $\phi(Z)$ denote the \gls{cpf} and \gls{pdf}, respectively. The equation uses the \gls{cpf} and \gls{pdf} to weigh the strength of either exploiting at the places already visited, or exploring, i.e., trying a new place with the highest uncertainty to see if an improvement on its prediction confidence can be achieved. The value of $\varphi_{EI}$ is then used to find $\mathbf{x}_i^{*}$ to evaluate $f$ next, which gives the current optimal $y_i^{*}$. 
    
Algorithm \ref{euclid2} summarises the proposed approach.\\
 
 \begin{algorithm}
 \caption{Bayesian optimisation sampling approach for saliency map generation}\label{euclid2}
 \begin{algorithmic}[1]
 \State \textbf{Start}: Random initialisation of $D_{1 : i} = \{\textbf{x}_{1 : i},y_{1 : i}\}$
 \For{iterations $i = 1,2,3,...,N$}{}
 \State Fit GP  model $g$ using $D_{1 : i}$ using Mat\'ern kernel
 \State Predict $m(Q)$ and covariance $k(Q,Q^{'})$ using $g$
 \State Evaluate $\varphi(Q)$
 \State Select $\mathbf{x}_i^{*} = \underset{\mathbf{x}}{\mathrm{arg\,min}} \,\varphi(Q)$
 \State Query $\mathbf{x}_{i}$ and then update $y_{i}$
 \State Augment dataset  $D_{1 : i+1} = D_{1 : i} \cup (\textbf{x}_{i},y_{i})$
 \State Repeat step $[3 - 8]$ until convergence or the $N^{th}$ iteration
 \EndFor
 \Return $m(Q)$ and covariance $k(Q,Q^{'})$
 \end{algorithmic}
 \end{algorithm}
 
 \subsection{Saliency map evaluation}\label{sub:eval}
In order to test the proposed saliency map generation method, and to benchmark it against the existing methods, a means of objectively evaluating the quality of the produced saliency maps is necessary. There exist a number of evaluation metrics for the saliency map approaches. In Kindermans et al. \cite{ref:Kindermans:2017}, an approach that assesses the reliability of pre-existing saliency methods was proposed. The method in~\cite{ref:Kindermans:2017} monitors the invariance of the saliency maps to small changes in the input, and reports the saliency method robustness, with the assumption that minor input changes that do not contribute towards a model's prediction should not alter the saliency map. Dabkowski and Gal~\cite{ref:Dabkowski:2017} proposed a saliency metric that evaluates the quality of the saliency map by finding the smallest rectangle containing the salient region that still allows for correct classification. This metric only considers saliency in terms of the final decision, and would thus penalise a saliency map that included an entire dog over one that simply included the dog's face, since a face is enough to perform classification. In order to address this limitation, we propose a saliency ratio metric $r_{sal}$, which requires that the bounding box of the desired object be known.

The saliency ratio metric is computed as the sum of the saliency score inside the bounding box over the total sum of the saliency map score. Let $\mathcal S$ represent the saliency map for the whole image, and $\mathcal T$ the saliency map for the target bounding box. The saliency ratio metric is given by:
\begin{equation}
    r_{sal} = \frac{\sum \mathcal T}{\sum \mathcal S}. 
    \label{label:forml1}
\end{equation}
This ratio metric evaluates saliency maps in a manner typically used for segmentation, and rewards saliency within the bounding box specified for a given target, while penalising saliency outside this region. The value of $r_{sal}$ is bounded to the $[0,1]$ interval, where $0$ corresponds to completely missing the bounding box in the saliency prediction, and $1$ corresponds to perfectly aligning the saliency prediction with the bounding box.
\section{Experimental setup and empirical results}\label{sec:experiments}
The experiments conducted for this study were designed to test the proposed method on a complex visual task, and to benchmark against the existing saliency map generation approaches.
The ImageNet 2012 dataset \cite{ref:russakovsky2015}, which mainly consists of different breeds of dogs, was used to evaluate the performance of the proposed approach. The model used for testing was the pre-trained VGG16 model \cite{ref:Simonyan:2014}, which is a high-performing CNN architecture. For all images tested, the saliency maps were generated with respect to the associated target class. For the Bayesian optimisation saliency overlay approach, 200 sample steps were used, where at each step a \gls{gp} model was fit using a set $\displaystyle{\mathcal{D} = \{\mathbf{x}_i, y_i\}_{i=1}^{200}}$ where $\displaystyle{\mathbf{x}_i}$ is made of $(u_i,v_i)$ image coordinates, and the blanking window size $\displaystyle{s_i \in S, S = \{50, 64, 78, 92, 107, 121, 135, 150\}}$. A constant colour 128 for the R, G, B channels was used for the blanked pixels. A total of 500 images were used to quantify the saliency map performance. A convergence analysis showed that 200 samples were typically enough for the optimisation to converge to a maximum on the Imagenet 2012 dataset.


\subsection{Saliency map generation}
\begin{figure*}[!tb]
    \centering
    \includegraphics[width=\textwidth]{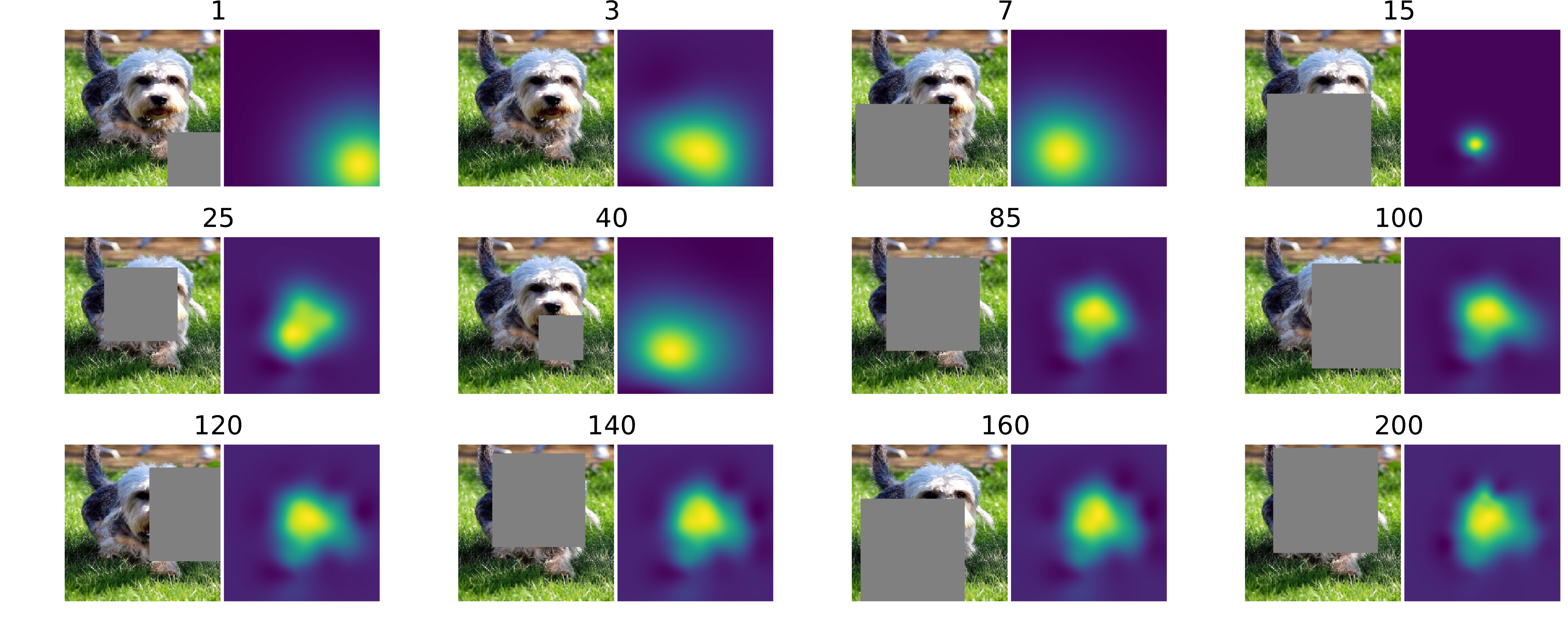}
    \caption{Illustration of the saliency overlay process using Bayesian optimisation method with a dog as the data sample. The figure shows pairwise input-output results obtained using 12 sample examples. The input images show varied blanking window sizes at different image locations. The output shows the saliency maps, where the blue region indicates pixel regions that are not important, and the green region indicates interesting pixels.}
    \label{fig:fig7}
\end{figure*}

The results shown in this section illustrate the selection criteria of the Bayesian optimisation approach for the saliency overlay generation. Figure \ref{fig:fig7} shows a subset of the input-output results, where all the input images are overlayed with a blanking window, and the corresponding output images show the saliency maps at a particular iteration. The blanked position and the blanking window size were specified by the \gls{ei} acquisition function. The predicted saliency overlays corresponding to the input images shown were produced using the predicted mean of the posterior \gls{gp}.

Figure \ref{fig:fig7} shows the automatic selection of the blanking window size, the sampling position, and the saliency maps at specific iteration times. The first saliency map is obtained using a \gls{gp} model fit with only one sample point, the second with three sample points, and the last with 200 sample points. It can be seen that as the sample size increases, the fitted \gls{gp} is able to produce a saliency map that can approximate the salient regions of the image using the black-box model, without evaluating the importance at every image pixel. From Figure~\ref{fig:fig7}, it can be seen that the saliency map gets better with an increase in sample points, and correctly indicates which image regions contributed more to the output of the given black box model. 

\subsection{Exploration-exploitation trade-off}
\label{results}

\begin{figure}[!tb]
    \centering
    \includegraphics[width=0.49\textwidth]{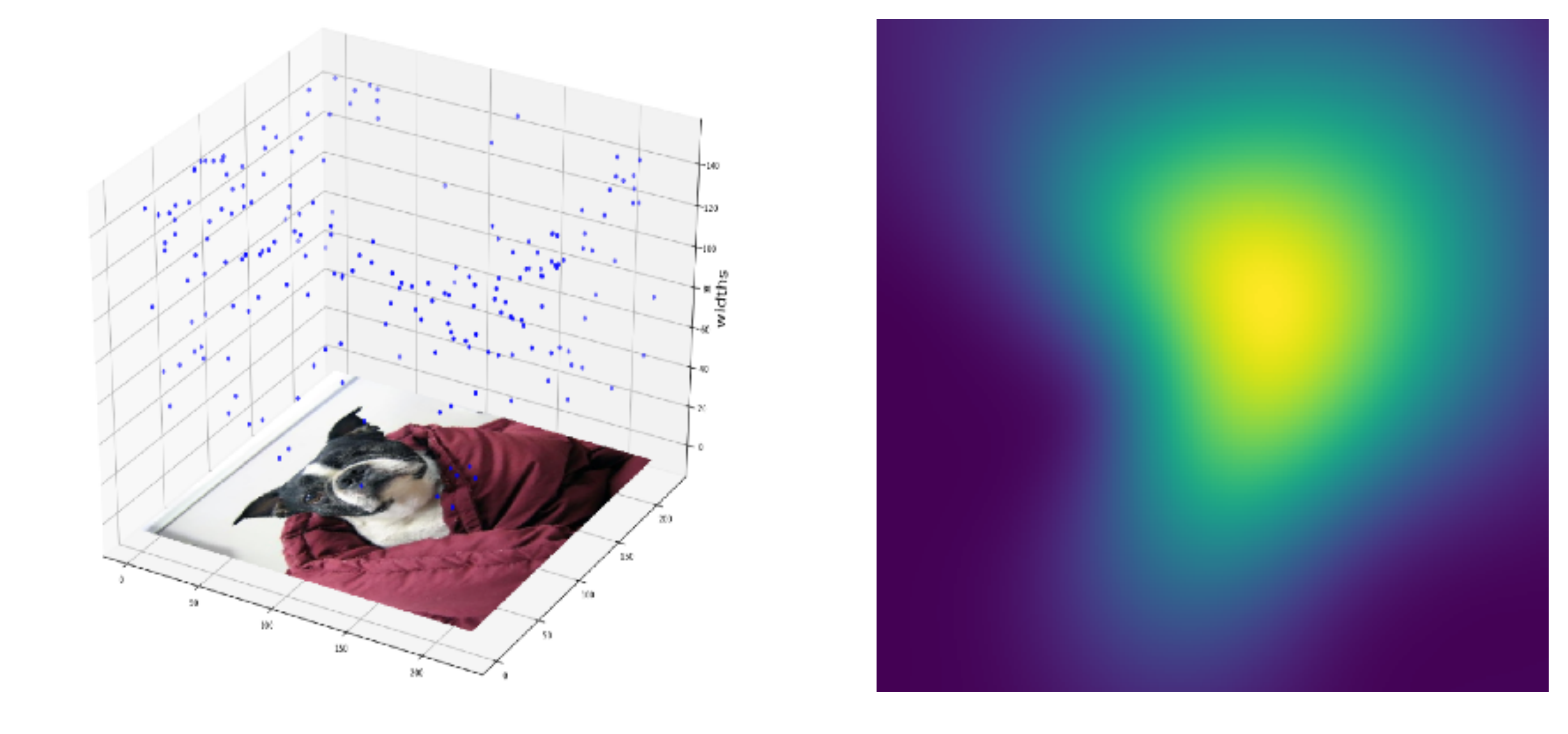}
    \caption{The saliency map and the points projection used for the saliency map generation.}
    \label{fig:fig5}
\end{figure}

The goal of the proposed saliency map technique is to generate the saliency map from the black-box model with the minimal number of samples, sampling substantially at regions of interest with various (or appropriate) window sizes, as opposed to the exhaustive search saliency map generation approach. Figure \ref{fig:fig5} shows a 3-dimensional diagram on the left illustrating the sampled points, and the corresponding saliency map to the right. The scattered points represent the $(x,y)$ image coordinates, and the blanking window sizes ($z$-axis) show the blanking values which were chosen to generate the saliency map shown on the right of Figure \ref{fig:fig5}. Certain image regions were blanked over more than other regions, where a part of the salient region (i.e. the dog's face) was also sampled substantially. Thus, the proposed technique exploits the area of interest, while still exploring in other regions of the image. With only 200 sample variables, the proposed Bayesian optimisation saliency overlay generation method was able to generate the saliency maps as shown in Figures~\ref{fig:fig7} and~\ref{fig:fig5}. The saliency maps produced show that the salient region was correctly identified.

In the next section, the proposed Bayesian optimisation method for saliency mapping is compared with other saliency mapping methods.

\subsection{Visual quality and interpretability of the saliency maps}

\begin{figure*}[!tb]
    \centering
    \includegraphics[width=\textwidth]{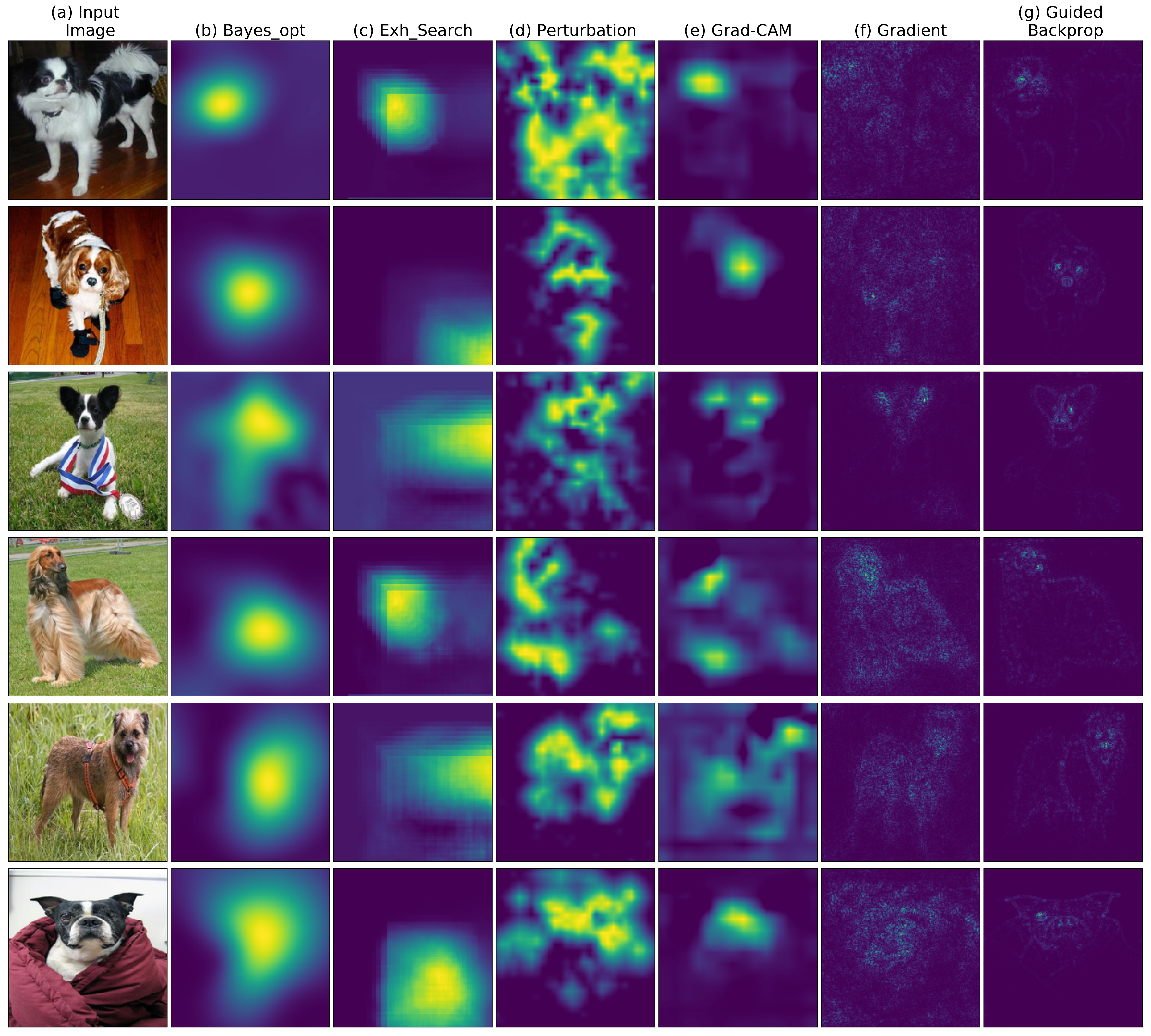}
	
	\caption{Saliency maps generated from the different methods as follows: (b) the proposed Bayesian optimisation method (Bayes\_opt), (c) the exhaustive search discussed in Section \ref{sec:occlusion}, (d) the saliency by perturbation \cite{ref:Fong:2018}, (e) the grad-CAM \cite{ref:Selvaraju:2016}, and (f) \cite{ref:Zeiler:2013}, (g) \cite{ref:Springenberg:2014} were obtained from \cite{ref:Maximilian:2018} toolbox.}
	\label{fig:fig2}
\end{figure*}
The proposed saliency mapping technique is a generalisation of the exhaustive search technique for saliency map generation, thus a comparison of the proposed technique to the exhaustive search technique is necessary. Figure~\ref{fig:fig2} presents a visual comparison of the saliency maps obtained from the proposed Bayesian optimisation approach, the exhaustive search approach, and other saliency map generation methods, applied to the pre-trained image classification black-box model~\cite{ref:Fong:2018, ref:Selvaraju:2016, ref:Simonyan:2013, ref:Springenberg:2014}. It is evident from Figure~\ref{fig:fig2} that the saliency maps from the proposed approach and the exhaustive search are comparable, producing saliency maps with a single prominent attention region. The saliency maps in Figure \ref{fig:fig2}~(d), produced using Fong's method \cite{ref:Fong:2018}, highlight multiple attention regions of very complex shapes, making such salient maps harder to interpret. 

By looking at the saliency maps obtained using the proposed method, it can be observed that the dog's head contributed the most as the attention region, since the salient regions are drawn over the region corresponding to the dog's head in the images. Further, the saliency maps produced by the proposed method are better than the saliency maps from the exhaustive search approach, since some of the attention regions on the exhaustive saliency maps (see Figure~\ref{fig:fig2} (b), rows 2, 3, and 6) do not correspond to the salient object of the input image. For example, the image in row two has the object of interest shown in the middle, but the saliency map from the exhaustive search shows the saliency region to be in the bottom-right corner of the saliency map.

Note that the proposed method and the exhaustive search method generate the saliency maps by simply sampling from a black-box model, with no prior information about the learned features. The saliency maps shown in Figure \ref{fig:fig2} illustrate that this simple technique is sufficient to capture the basic saliency information. The saliency maps produced by the proposed method successfully highlight the interesting/important regions that the model has used to make the class prediction. More complex methods illustrated in Figure~\ref{fig:fig2} (d) - (g) optimise the blanking mask or the gradients that change the classification result of a model. In contrast, the proposed approach and the exhaustive search approach do not require any information about the model parameters or the object of interest from the learned features to produce the saliency map.
\subsection{Localisation of the saliency map}
\label{label:ratio_formula}
\begin{figure}[!b]
    \centering
    \includegraphics[width=0.48\textwidth]{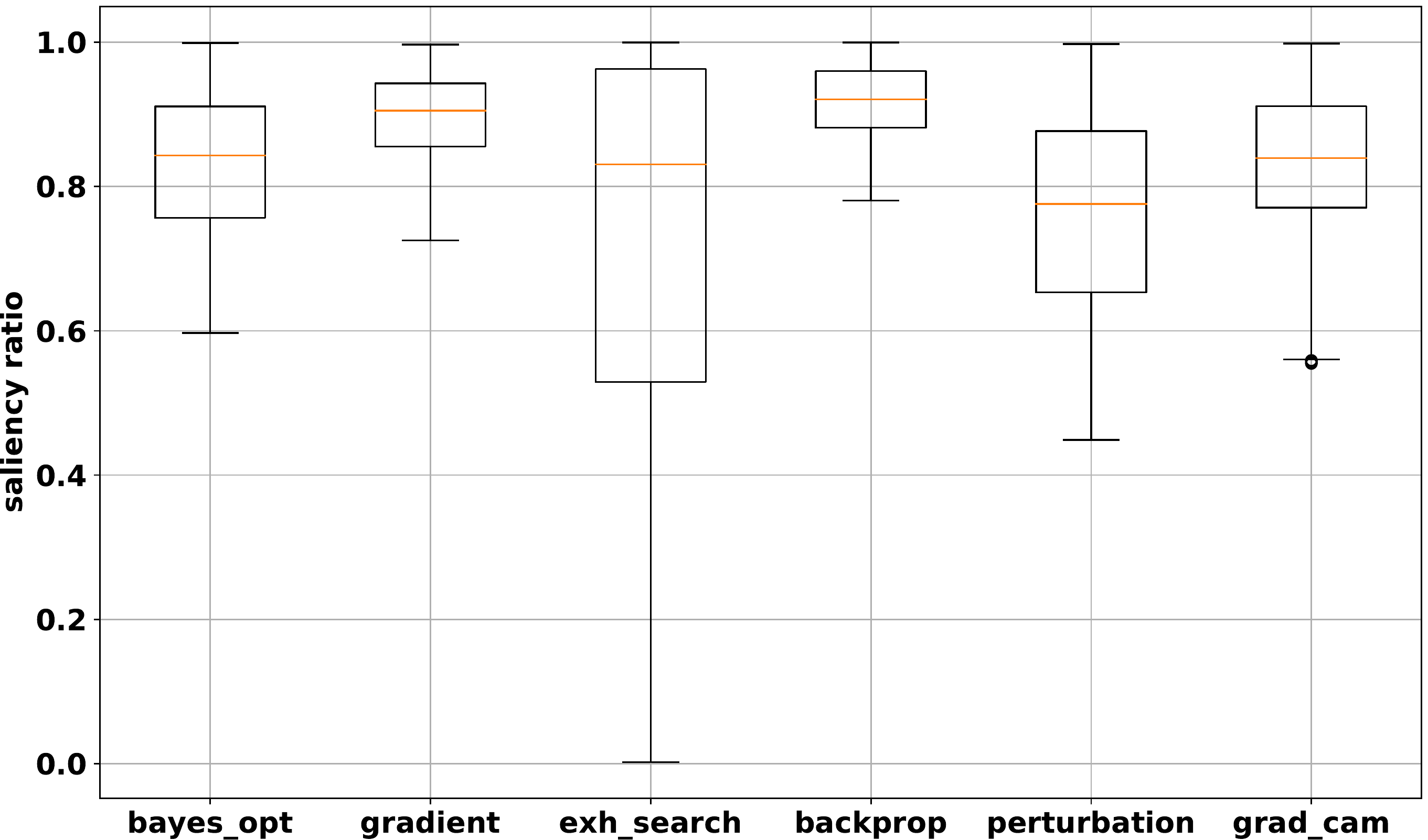}
    \caption{A comparison of the various saliency mapping methods using the proposed saliency ratio metric, $r_{sal}$. Same naming abbreviations as in Figure~\ref{fig:fig2} are used for the algorithm names.}
       \label{fig:fig3}
        \end{figure}

The saliency maps were generated over $500$ images, and the saliency ratio metric defined in Section~\ref{sub:eval} was computed for each map. The results are shown in Figure~\ref{fig:fig3} for the proposed Bayesian optimisation approach, the exhaustive search, and four other approaches that require access to model parameters. For each bar in Figure~\ref{fig:fig3}, an overview of how the saliency overlay algorithm performed is shown using the statistical five number summary, namely: the minimum, the first quartile, the mean, the third quartile, and the maximum, together with the outliers represented by circular points below the minimum. In Figure~\ref{fig:fig3}, the $r_{sal}$ values of 1 represent perfect saliency maps, and $r_{sal}$ values of $0$ correspond to saliency maps that fall outside the target bounding box. The results in Figure~\ref{fig:fig3} show that the proposed Bayesian optimisation saliency method outperformed the exhaustive search and the perturbation method, and was comparable with the gradient-aware methods~\cite{ref:Fong:2018, ref:Selvaraju:2016,ref:Springenberg:2014, ref:Simonyan:2013}. The exhaustive search saliency method performed the worst as compared to the other saliency map algorithms. Thus, the Bayesian optimisation approach is a viable technique for saliency map generation, especially when one cannot get access to the model's parameters.

The computational efficiency of the Bayesian optimisation approach is discussed in the next section.

\subsection{Time complexity}
The computational complexity of generating the saliency maps using the Bayesian optimisation approach increases cubically with an increase in the number of training variables. The reason behind this is that \gls{gp} models are computationally expensive, requiring more time as more data is used to fit the model. However, the proposed Bayesian optimisation method generates the saliency map before the fitting of \gls{gp} reaches the pick cubic computational time, making it computationally feasible.

A comparison of the time efficiency of the exhaustive approach and the Bayesian optimisation approach is presented in Table \ref{tab3:tab1}. Table \ref{tab3:tab1} shows the time complexity of the saliency map generation, where $m,n,l$ represent the number of the $x$ image values, the  $y$ image values and the $s$ blanking window values, respectively, and $N$ is the number of model iterations. The $\bigO(\boldsymbol{\cdot})$ notation indicates the number of model evaluations required when performing the experiment. With the exhaustive search approach, time increases linearly with the increase in the number of sample evaluations required to generate the saliency map. For the Bayesian optimisation approach, the time required to run the algorithm increases cubically with an increases in the sample variables. The results presented in Table \ref{tab3:tab1} show that for larger images, it takes the exhaustive approach more time to generate the saliency map. Therefore, the Bayesian optimisation method requires fewer samples and uses less computational time than the exhaustive search. Thus, the Bayesain optimisation method for saliency map generation is more computationally efficient. 

\begin{table}[!h]
    \centering
    \caption[Computational complexity table]{Computational complexity}
    \resizebox{0.48\textwidth}{!}{
    \begin{tabular}{|c|c|c|}
    \hline
         Approach & Time & Num. observations\\
                 & complexity & (ImageNet)\\
         \hline \hline
         Exhaustive & $\bigO(m\times n \times l)$ & 401 408\\
         Bayesian opt. & $\bigO(N^3)$ & 200 \\
         \hline
    \end{tabular}}
    
    \label{tab3:tab1}
\end{table}

\section{Conclusion}\label{sec:conclusion}
In this study, we proposed a method for the generation of saliency maps from black-box models using a Bayesian optimisation sampling approach,  generalising an occlusion-based sliding window approach. Instead of sliding a range of blanking windows exhaustively over pixel locations, the proposed approach selects occlusion positions and blanking region sizes based on previous perturbation results. The results obtained show that this approach produces improved saliency maps, and performs similarly to gradient-based approaches, which require access to model parameters. This means that the proposed approach is a useful interpretability aid in black-box model situations, where model introspection is required without parameter access.

In future, additional criteria can be added for blanking. This includes adding variables such as height and width blanking parameters, or different shapes for the blanking window such as triangular or ellipsoid. Selecting the colour to use, as well as choosing the alternative methods to fill the blanking window such as blurring the image, could also be explored. In this work, the \gls{gp} is used as a proxy model, which requires computing $N \times N$ covariance function manipulations for $N$ observations, making it computationally expensive as $N$ increases. Thus, alternative proxy models such as sparse \gls{gp} or other approaches could be evaluated as a substitute. 
{\small
\IEEEtriggeratref{11}

\newcommand{\noopsort}[1]{} \newcommand{\singleletter}[1]{#1}

}

\end{document}